\documentclass[11pt,a4paper]{article}
\usepackage[hyperref]{emnlp2020}
\usepackage{times}
\usepackage{latexsym}

\usepackage{microtype}
\usepackage{graphicx}
\usepackage{amsmath,amsfonts,amstext,amssymb,amsthm}
\usepackage{multirow}
\usepackage{array}
\usepackage{subcaption}
\usepackage{comment}
\usepackage{dirtytalk}

\newcolumntype{L}[1]{>{\raggedright\let\newline\\\arraybackslash\hspace{0pt}}m{#1}}
\newcolumntype{C}[1]{>{\centering\let\newline\\\arraybackslash\hspace{0pt}}m{#1}}
\newcolumntype{R}[1]{>{\raggedleft\let\newline\\\arraybackslash\hspace{0pt}}m{#1}}

\aclfinalcopy % Uncomment this line for the final submission
\title{An efficient representation of chronological events in medical texts}

\author{Andrey Kormilitzin\textsuperscript{1}\footnotemark[1], Nemanja Vaci\textsuperscript{2}\footnotemark[1], Qiang Liu\textsuperscript{1}, Hao Ni\textsuperscript{3,5}, 
\\ {\bf Goran Nenadic\textsuperscript{4,5}}  \and {\bf Alejo Nevado-Holgado\textsuperscript{1,6}} \\
\textsuperscript{1}Department of Psychiatry, University of Oxford, Oxford, OX3 7JX, UK \\ 
\textsuperscript{2}Department of Psychology, University of Sheffield, Sheffield, S1 1HD, UK \\
\textsuperscript{3}Department of Mathematics, University College London, London, WC1H 0AY, UK \\
\textsuperscript{4}School of Computer Science, University of Manchester, Manchester, M13 9PL, UK \\
\textsuperscript{5}The Alan Turing Institute London, London, NW1 2DB, UK \\
\textsuperscript{6}Akrivia Health, Oxford, OX1 1BY, UK\\
\tt andrey.kormilitzin@psych.ox.ac.uk}

\date{}

\begin{document}
\maketitle
\footnotetext[1]{Equal contribution.}
\begin{abstract}
In this work we addressed the problem of capturing sequential information contained in longitudinal electronic health records (EHRs). Clinical notes, which is a particular type of EHR data, are a rich source of information and practitioners often develop clever solutions how to maximise the sequential information contained in free-texts. We proposed a systematic methodology for learning from chronological events available in clinical notes. The proposed methodological {\it path signature} framework creates a non-parametric hierarchical representation of sequential events of any type and can be used as features for downstream statistical learning tasks. The methodology was developed and externally validated using the largest in the UK secondary care mental health EHR data on a specific task of predicting survival risk of patients diagnosed with Alzheimer's disease. The signature-based model was compared to a common survival random forest model. Our results showed a 15.4$\%$ increase of risk prediction AUC at the time point of 20 months after the first admission to a specialist memory clinic and the signature method outperformed the baseline mixed-effects model by 13.2 $\%$.

\end{abstract}

%We introduced a novel methodology for maximising predictive information from the longitudinal patients' trajectories of ordered events recorded in electronic health records.
%real-world observational electronic patients data from secondary care UK Mental Health NHS Trusts were used \citep{goodday2020maximizing}.For the epidemiological survival analysis real-world observational electronic patients data from secondary care UK Mental Health NHS Trusts were used \citep{goodday2020maximizing}. 

\section{Introduction}
% recognised the importance of the order of events contained in EHRs to make more accurate predictions and
% (e.g. focusing on patients that receive monotherapy treatment).
Electronic health records (EHRs) have now become ubiquitous and offer novel opportunities for clinical research by supporting the development of intelligent decision support systems and improvement of patients' care. One of the distinct features of EHR is that the data are being collected over time and might be seen as health data streams, allowing research to study longitudinal trends and make inference about the progression of disease, treatments and outcomes. However, the proper representation of sequential medical events still remains a challenge. Moreover, longitudinal clinical notes exhibit a multi-level hierarchical structure, where events are described and embedded in sentences, sentences in paragraphs and eventually resulting in chronologically ordered documents. Recent works have addressed the problem of capturing this information directly from raw texts by introducing novel neural network architectures, such as attention-based recurrent neural networks \cite{bai2018interpretable} and time-aware Transformers \cite{zhang2020time}. When dealing with chronological clinical notes, practitioners make multiple decisions on how to structure and transform these sequential events, which are often simplifications of medical histories. In this work we proposed a different methodology to address the problem of learning from events found in clinical notes, by first extracting them using natural language processing and then representing the sequential order by means of the {\it path signatures}. The signature \cite{lyons2014rough} is a non-parametric representation of heterogeneous sequential data, offers a feature extraction method from longitudinal events and can naturally be integrated within a general data mining pipeline. To demonstrate the methodology, we used the largest secondary care mental health EHR data in the UK to develop a survival prognostic model for patients diagnosed with Alzheimer's disease. 

%Using the information extraction system, the free-text medical records were parsed into a structured format, comprising the longitudinal information about trialled medications, time intervals and cognitive health. We demonstrated that the signature-based prognostic model was able to learn various treatment pathways of switching medications and the trajectory of cognitive health over time and outperformed a standard statistical approach of survival modelling. 
%In this work we developed a conceptual approach to modelling longitudinal events extracted from free-text clinical records.

\section{Method}

\subsection{Data}

The data in this study were sourced from the UK-Clinical Record Interactive Search system (UK-CRIS), which provides a research platform (https://crisnetwork.co/) for data mining and analysis using de-identified real-world observational electronic patients records from twelve secondary care UK Mental Health NHS Trusts \citep{goodday2020maximizing}. UK-CRIS provides access to structured information, such as ICD-10 coded diagnoses, quality of life scales and demographic information, as well as various unstructured texts, such as clinical summaries, discharge letters and progress notes. The study cohort jointly comprised records from 24,108 patients diagnosed with Alzheimer's disease and various types of dementia, containing more than 3.7 million individual clinical documents from two centres: Oxford and Southern Health Foundation NHS Trusts. The field of clinical NLP in general, and of mental health and Alzheimer's research in particular, largely suffers from the dearth of gold-annotated data. The reason is due to the shortage of trained annotators with clinical background who are also authorised to access sensitive patient-level data. Therefore, to develop a robust information extraction (IE) model from an insufficient amount of data, we leveraged the idea of transfer learning using the publicly available MIMIC-III corpus \citep{johnson2016mimic} comprising information relating to patients admitted to intensive care units (ICU) with more than 2.1 million clinical notes as well as 505 gold-annotated by clinical experts discharge summaries from the 2018 n2c2 challenge \citep{henry20202018}. We assert that the study was independently approved and granted by the Oxfordshire and Southern Health NHS Foundation Trust Research Ethics Committees. %Individual patient consent was not required for the use of anonymised data.  

\subsection{Information extraction model}

The information extraction model was developed to identify diagnosis, medications and cognitive health assessment Mini-Mental State Examination score (MMSE) \cite{pangman2000examination}. Additionally, the identified entities were classified according to several attributes, such as the 'experiencer' modality (i.e., whether the MMSE was actually referring to a patient or to a family member), temporal information (i.e the date of diagnosis or MMSE score) and negations (i.e. discontinued medications) \citep{harkema2009context, gligic2019named}. Such drug mentions were discarded in order to extract the most accurate information. Generic and brand drug names were normalised using the British National Formulary, the core pharmaceutical reference book \citep{joint2019bnf}. The architecture of the named entity recognition model comprised a hybrid approach of an ontology-based fuzzy pattern matching and a bi-directional LSTM neural network architecture with the attention mechanism \citep{bahdanau2014neural} for sequence classification. The GloVE word embedding \cite{pennington2014glove} were fine-tuned on both MIMIC-III and UK-CRIS data \citep{vaci_koychev_kim_kormilitzin_liu_lucas_dehghan_nenadic_nevado-holgado_2020, kormilitzin2020med7}. The developed IE model was trained only on data from the Oxford Health NHS Trust instance and externally validated on a sample of data from a regionally different Southern Health NHS Foundation Trust.

\subsection{The signature of a path}
\label{sec:signatures}

\begin{figure}[htp!]
    \centering
    \includegraphics[width=0.4\textwidth, height=3.5cm]{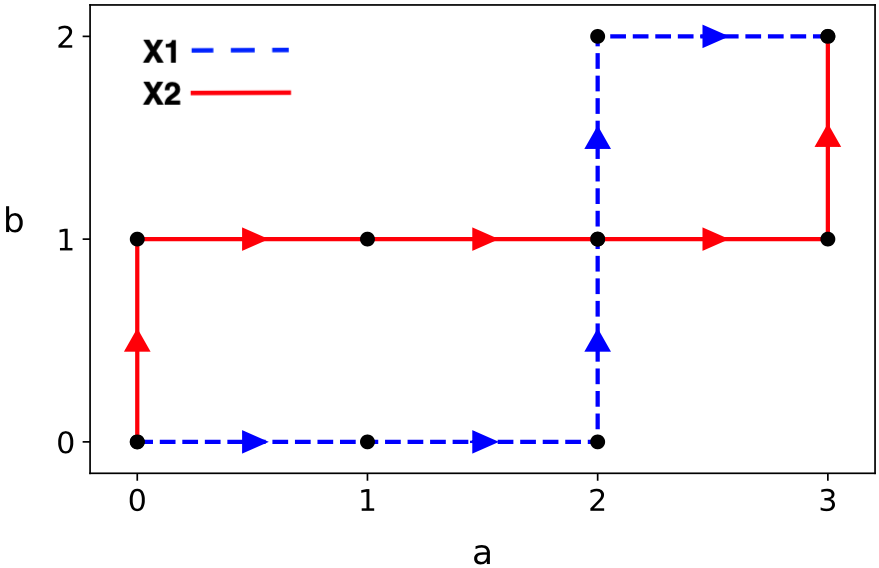}
    \caption{Two paths $X_1$ = {\it aabba} and $X_2$={\it baaab}.}
    \label{fig:two_paths}
    \vspace{-0.0cm}
\end{figure}

Repeated measurements, speech, text, time-series or any other sequential data might be seen as a path-valued random variable. Formally, a path $X$ of finite length in \textit{d} dimensions can be described by the mapping $X:[a, b]\rightarrow\mathbb{R}$ $\!\!^d$, or in terms of co-ordinates $X=(X^1_t, X^2_t, ...,X^d_t)$,  where each coordinate $X^i_t$ is real-valued and parametrised by $t\in[a,b]$. The signature representation $S$ of a path $X$ is defined as an infinite series:
\noindent
\begin{equation}\label{eq:path_signature}
    \begin{split}
        S(X)_{a, b} = (1, & S(X)_{a, b}^1, S(X)_{a, b}^2, ..., S(X)_{a, b}^d, \\
                          & S(X)_{a,b}^{1, 1}, S(X)_{a,b}^{1, 2}, ...),
    \end{split}
\end{equation}
\noindent
where each term is a $k$-fold iterated integral of the path $X$ labelled by multi-index $i_1,...,i_k$:
\begin{equation}\label{eq:sig_moments}
    S(X)_{a, b}^{i_1,...,i_k} = \int_{a<t_k<b}...\int_{a<t_1<t_2} \mathrm{d}X_{t_1}^{i_1}...\mathrm{d}X_{t_k}^{i_k}.
\end{equation}
\noindent %%and one-hot-encoded in co-ordinates of (a,b) as presented in Table \ref{tab:one_hot_encoding_of_paths}. 
However, in many real-life applications the first $k$-terms of the truncated signature at level $L$ give a satisfying approximation. Intuitively, it is analogous to statistical moments of a $d$-dimensional vector-valued random variable, such as mean, variance or higher moments. One can define statistical moments of a \textit{path}-valued random variable, which are essentially the {\it signature} moments \citep{chevyrev2018signature} defined in Eq. \eqref{eq:sig_moments}. The signature $S(X)$ completely characterises a path $X$ up to tree-like equivalence and is invariant to reparameterisation \citep{hambly2010uniqueness}. The signature can also be expressed in a more compact form known as {\it log-signature} \cite{liao2019learning, morrill2020neural}, which is the formal power series of $\log S(X)$, while carrying the same information. Informally, the path signature captures the order of events. For example, consider two sequences $X_1$ = {\it aabba} and $X_2$ = {\it baaab} consisting of a simple vocabulary with only two letters $\{a, b\}$. The sequences might be presented as paths in 2d space as shown in Fig. \ref{fig:two_paths}. 
\noindent
Each linear segment between two points (Fig. \ref{fig:two_paths}) corresponds to a single letter in the sequence and the arrows denote the temporal direction of the sequence. 
\begin{table}[h]
    \centering
    \resizebox{\columnwidth}{!}{
    \begin{tabular}{c|cc|c|cc|ccc}
        Level  & \multicolumn{2}{c|}{1} & \multicolumn{1}{c|}{2} & \multicolumn{2}{c|}{3} & \multicolumn{3}{c}{4} \\\hline\hline
        $S(X_1)$  & 3  &  2   &  {\bf 1}   & {\bf -0.5}  & {\bf -1}   & {\bf -1/3} & {\bf -0.5}  &  0 \\
        $S(X_2)$  & 3  &  2   &  {\bf 0}   & {\bf  1.5}  & {\bf 0.5}  & {\bf    0} & {\bf 0} & 0 \\ \hline
    \end{tabular}}
    \caption{The first $k=8$ terms of the log signature expansion up to level $L=4$. The difference between two sequences $X_1$ and $X_2$ is apparent starting from the second level.}
    \label{tab:log_sig_of_paths}
\end{table}
\noindent
\vspace{-0.2cm}
Despite the same number of letters in the sequences $\{a=3, b=2 \}$, the order of letters matters. The signature easily picks the differences and the first four levels of the log-signatures of paths are shown in Table \ref{tab:log_sig_of_paths}. The lower order signature terms $S^{(i)}$ are the increments along the $i$-th direction (i.e. the distance between the endpoints), for example, $S^{(1)}=3-0=3$ and $S^{(2)}=2-0=2$ as can be seen in Figure \ref{fig:two_paths}. The second order corresponds to the area enclosed by a path and a chord connecting endpoints \cite{chevyrev2016primer}. 

The usefulness of a path signature as a feature map of sequential data was demonstrated theoretically \citep{chevyrev2018signature} as well as in numerous machine learning applications in healthcare \citep{morrill2019signature, kormilitzin2016application, arribas2017signature,  morrill2020utilization, kormilitzin2017detecting}, finance \citep{arribas2018derivatives}, computer vision \citep{yang2017leveraging, xie2017learning}, topological data analysis \citep{chevyrev2018persistence} and deep learning \citep{kidger2019deep}.
\begin{table*}[t]
    \centering
%    \subfloat[]{
    \resizebox{2\columnwidth}{!}{
    \begin{tabular}{L{2.2cm} L{16cm}}
        Doc date    & Text \\\hline\hline
        05-Oct-2016 & {\it Today I saw a patient diagnosed with Alzheimer's, who deteriorated: MMSE 23/30 as compared to 25/30 from 1st January. Started on Rivastigmine.}\\
        12-Feb-2017 & {\it Today MMSE 19, the patient didn't respond to Rivastigmine and was changed to Donepezil.}         \\
        03-Feb-2018 & {\it Great response to new treatment (MMSE 23/30), continue on Donepezil.} \\
        01-Apr-2019 & {\it The patient stopped responding to Donepezil and severely deteriorated (MMSE 14/30), stop Donepezil.} \\\hline
    \end{tabular}}%}
    \caption{A synthetic example of chronological medical records.}
    \label{tab:records}
\end{table*}
\noindent
\vspace{-0.0cm}
\noindent

\subsection{Independent and outcome variables}

The independent variables used in the prognostic model were medications and the MMSE scores collected over time. The dependent outcome variable was right-censored time to death data in months. A synthetic example of the patient's records (Table \ref{tab:records}) and the corresponding algorithmically extracted longitudinal data is presented in Table \ref{tab:longitudinal_data_example}. The outcome variable was encoded as a tuple: $(True, 34.17)$ indicating that a person has died after 34.17 months since the very first visit to a specialist memory clinic. The patient was treated by two different medications with a changing pattern and eventually was tapered off medication due to no further expected improvement.

\begin{table}[htp!]
    \centering
    \resizebox{0.99\columnwidth}{!}{
        \begin{tabular}{L{3cm}L{3cm}L{1.5cm}}
        Date        &   Medication      & MMSE    \\\hline\hline
        01-Jan-2016  &   NoMed           & 25/30 \\
        05-Oct-2016  &   Rivastigmine    & 22/30 \\
        12-Feb-2017  &   Donepezil       & 19/30 \\
        03-Feb-2018  &   Donepezil       & 23/30 \\
        01-Apr-2019  &   Discontinued    & 14/30 \\\hline
    \end{tabular}}
    \caption{Extracted and chronologically structured data from Table \ref{tab:records}.}
    \label{tab:longitudinal_data_example}
\end{table}
\vspace{-0.0cm}
\noindent
%including dates of medical encounters, medications and the MMSE scores. \say{NoMed} denotes that a patient was not prescribed a drug at the first visit to a memory clinic and \say{Discontinued} to stop medical treatments
\subsection{Baseline longitudinal data summarisation}
\label{sec:linear_features}
The signature transformation might be seen as a hierarchical statistical summarisation (\say{feature extraction}) of the longitudinal data along the temporal dimension. In order to benchmark the proposed method, we used a time-honoured linear mixed-effects regression as a baseline model for longitudinal summarisation. Specifically, each patient-level longitudinal MMSE scores were modelled using a linear regression and the resulting coefficients, such as an intercept and a slope, were used as features representing the progression of the MMSE over time. The median number of medication categories was used as an additional feature, resulting in three features for each patient.

\subsection{Survival random forests}

The common statistical approach to analyse the time-to-event survival data is based on the linear Cox model \cite{collett2015modelling}. However, \citet{miao2015random} showed that a survival random forest (SRF) approach \cite{ishwaran2008random} outperformed linear Cox model, based on the Harrell’s concordance index (C-index) \cite{harrell1982evaluating}, and was understandably capable of identifying non-linear effects of the input variables as opposed to linear Cox model. Therefore, we chose the SRF as the preferred method. The SRF approach was implemented in Python using \say{scikit-survival} package \cite{polsterl2015fast}. The Harell's C-index (the concordance index) is a goodness of fit measure for risk scores models. It is a common statistical approach to evaluate risk models in survival analysis, where data may be right-censored and corresponds to rank correlation between predicted risk scores and observed time points, similarly to Kendall’s $\tau$.

\vspace{-0.0cm}
\section{Results}

\subsection{Information extraction model}
%Additionally, to alleviate the problem of imbalanced distribution of concepts, we created a synthetic 'silver' corpus using rule-based pattern matching with fine-tuned \say{sense2vec} embedding \cite{trask2015sense2vec} on UK-CRIS data. 
We used a hybrid approach to developing an IE model consisting of training a baseline model using MIMIC-III and n2c2 annotated data. Specifically, the named-entity recognition (NER) model comprised a transition-based system based on the chunking model \cite{lample2016neural} where tokens were represented as hashed and embedded representations of the prefix, suffix, shape and lemmatised features of words, followed by the rule-based matching using the BNF vocabulary. The IE model was implemented using \say{spaCy} python library\footnote{https://spacy.io}, including negations and temporal information identification as well as relationships classification between the word-tokens using linguistic features, such as part-of-speech and dependencies. Finally, the active learning tool \say{Prodigy}\footnote{https://prodi.gy} was used for iterative model improvement \cite{vaci2020natural}. Target domain training, validation and external validation data contained a collection of gold-annotated drug names, diagnosis and cognitive health assessment MMSE scores as shown in Table \ref{tab:annotated_summary_stats}.
\begin{table}[h]
    \centering
    \resizebox{\columnwidth}{!}{
    \begin{tabular}{lrrrr}
        Concept  &   Training    &   Validation  &   External val.       & Total   \\ \hline\hline
        Drug                &   216       &       153   &                   30  & 399   \\
        MMSE                &   169         &       87      &                   23  &  279 \\
        Diagnosis           &   570         &       352     &                   26  &  948 \\\hline
%        Num. of docs   &   360         &       240     &                   20  & 620   \\\hline
    \end{tabular}
    }
    \caption{The number of gold-annotated instances in the training, validation and external validation data sets.}
    \label{tab:annotated_summary_stats}
    \vspace{-0.3cm}
\end{table}
\noindent
The IE model achieved a good and consistent performance on both validation and external validation data sets (Table \ref{tab:ie_results}). The annotation schema was developed following the recommendations of \citet{pustejovsky2012natural}. The token-level performance metrics were evaluated using the SemEval schema \cite{segura2013semeval} and the inter-annotator agreement (IAA) of two clinical annotators was computed using F1 score.

\begin{table}[htp!]
    \centering
    \resizebox{\columnwidth}{!}{
    \begin{tabular}{llll|lll|l}
                    &   \multicolumn{3}{c|}{Validation} & \multicolumn{3}{c|}{External val.}& IAA   \\
        Concept     &   Pr      &   Re      &   F1      &    Pr     &  Re   &   F1     &   F1      \\ \hline\hline
        Diagnosis   &   89.6    &   96.3    &   92.8    &    84.1   &  86.3 &   84.8      & 95            \\
        Drug        &   98      &   98      &   98.1    &    92.4   &  68.4 &   78.3      & 96           \\
        MMSE        &   92.6    &   74.7    &   82.8    &    85.6   &  81.2 &   82.6      & 100          \\\hline
%        Experiencer &   89      &   91.1    &   90  &           & 100           \\
%        Negation    &   88.7    &   92.1    &   90.3&           & 96              \\
%        Date info   &   90.8    &   91.9    &   91.2&           & 90          \\\hline    
    \end{tabular}
    }
    \caption{Performance (shown in $\%$) of the information extraction model. IAA - inter annotator agreement.}
    \label{tab:ie_results}
\end{table}
\noindent
\vspace{-0.7cm}
\begin{table}[htp!]
    \centering
    \resizebox{\columnwidth}{!}{    
    \begin{tabular}{ccccc}
                    &       n   &   male        & female  & survival time   \\\hline\hline
        died        &   1962    &   841         & 1121    & 52.2(22.8)      \\
        censored    &   1500    &   529         & 971     & 28.4(16.6)      \\\hline
    \end{tabular}}
    \caption{Summary statistics of the extracted data for survival analysis. Survival time is shown as mean(std) in months. The MMSE scores were not observed for censored people later in time, while date of death was recorded in hospital.}
    \label{tab:my_label}
\end{table}

\vspace{-0.5cm}
\subsection{Prognostic model}

Four prognostic models were developed and compared to each other. All models estimated the survival probability of a patient diagnosed with Alzheimer's disease since their first admission to a memory clinic. We compared signature (\say{Sig}, Sec. \ref{sec:signatures}) versus non-signature (\say{Non-sig}, Sec. \ref{sec:linear_features}) models. We also estimated the added value of the sequential information contained in the treatment course with medications. Specifically, we used two sets of input variables: $\{$time, MMSE$\}$ and $\{$time, MMSE, medications$\}$, where time corresponds to the date of MMSE score or prescribed medication as presented in Table \ref{tab:longitudinal_data_example}. For the \say{Sig} model, the input variable were first transformed into signatures, where the categorical medication names were one-hot encoded and augmented with numerical MMSE scores to create a path. For the \say{Non-sig} model, the longitudinal MMSE scores were summarised by means of linear models adjusting for each patients and the median number of distinct medications were computed. Both models were trained and validated using the same folds of stratified 5-fold cross validation (with fixed random seed). The quality of predictions was assessed using the Harell's C-index and the results are summarised in Table \ref{tab:results}. The signatures were computed using the \say{esig} Python library\footnote{https://esig.readthedocs.io/}, however, alternative libraries are also available \cite{reizenstein2018iisignature, kidger2020signatory}.

\begin{figure}[htp!]
    \centering
    \includegraphics[width=0.48\textwidth, height=5cm]{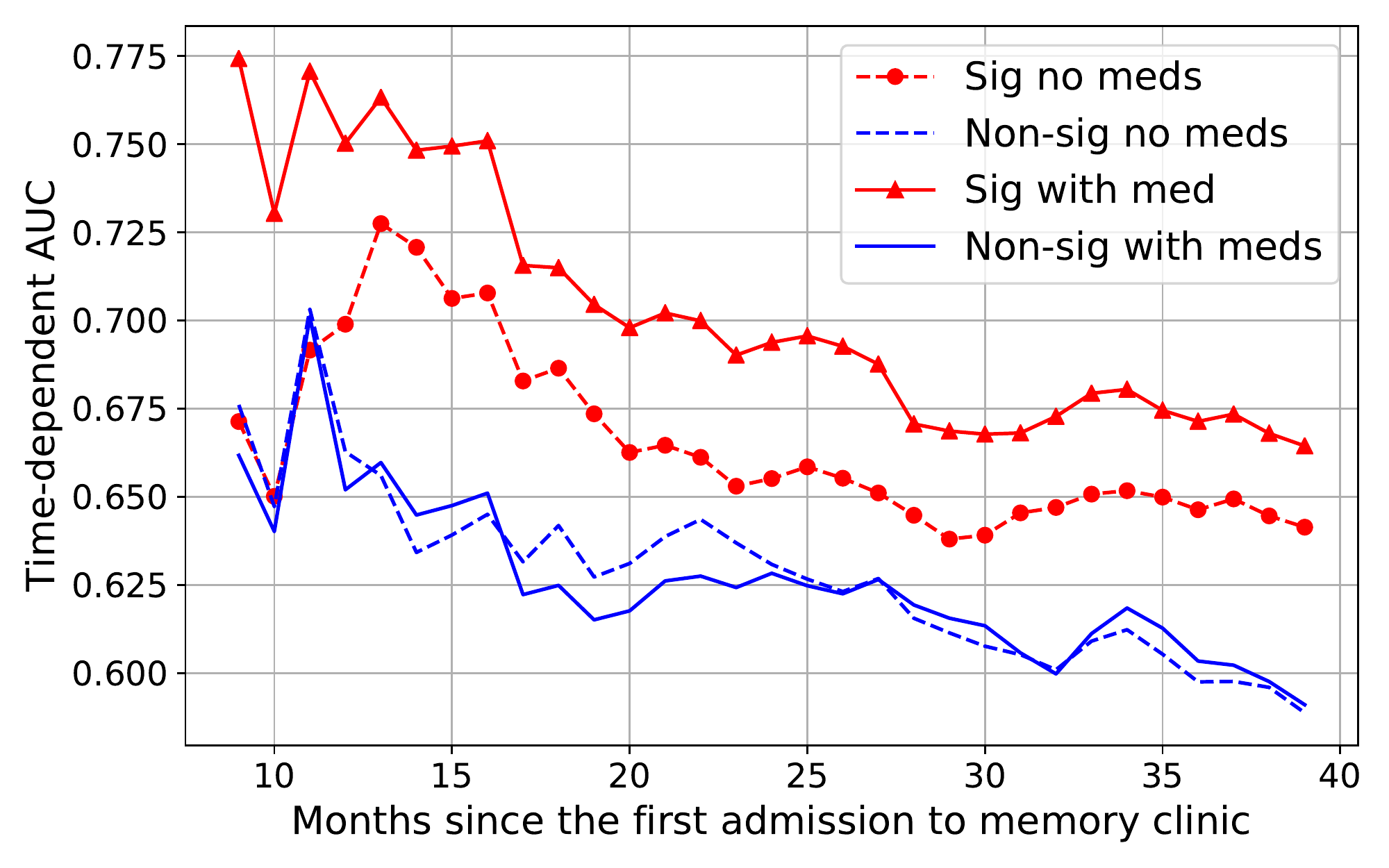}
    \caption{Time-dependent AUC of risk prediction over time since the first admission to a memory clinic.}
    \label{fig:auc}
\end{figure}

\begin{table}[htp!]
    \centering
    \resizebox{\columnwidth}{!}{
    \begin{tabular}{lccc}
     Features       &   Sig   &   Non-sig   \\\hline\hline
     $\{$time, MMSE$\}$   &   0.626(0.009)        &   0.574(0.022)      \\
     $\{$time, MMSE, meds$\}$   &   0.621(0.011)        &   0.571(0.019) \\ \hline
    \end{tabular}}
    \caption{Harell's C-index measure of four models. Values reported as mean(std) over 5-fold cross validation.}
    \label{tab:results}
\end{table}
\noindent
We also estimated the time-dependent area under the curve of receiver operating characteristics \cite{lambert2016summary}. It is a natural extension of a common AUC ROC analysis to possibly censored survival times where the patients' cognitive health is usually better at the very first visit to a memory clinic, while their condition may deteriorate later. The time-dependent cumulative dynamic AUC ROC of all four models are presented in Fig. \ref{fig:auc}. The signature features outperformed the non-signature ones at all times and the inclusion of sequential information from switching medications improved AUC ROC at later times. However, both models struggle to reliably predict the future outcomes further than 3 years. This is due to the limitation of predictors and the available number of patients after 3 years rather than the capacity of our model.

\vspace{-0.0cm}
\section{Discussion and future direction}
Unstructured longitudinal electronic health records, such as free-text clinical notes, inherently contain rich information about patients' health and outcomes over time. The right analytical tools capable of capturing sequential information can therefore maximise utilisation of longitudinal EHRs and can be valuable for supporting clinical decisions and prognostic models. In this work we implemented a signature-based approach to represent chronological events extracted using natural language processing from clinical notes. Extracted chronological events can be seen as a trajectory (path) embedded in a high-dimensional multi-modal space of events (i.e. different medications, interventions, measures, etc) and the signature uniquely characterises the path in the most succinct way. The signature-based feature extraction approach was compared to hand-crafted features, comprising a slope and an intercept of MMSE scores over time and the median number of medications for each patient. The signatures represent a hierarchical collection of features, where the first order is proportional to linear statistical moments (i.e. mean) and is not sensitive to the order of data points, as illustrated in Table \ref{tab:log_sig_of_paths}. We demonstrated that the sequential information about medications has significantly improved the time-demented AUC as captured by the signatures (Figure \ref{fig:auc}). In future works we will extend the proposed framework to include the structured information available in EHR (i.e. lab results, coded procedures or clinical encounters) and will develop an interpretability framework to make the signature-based models explainable. 
%The data used in the study can be accessed using the UK-CRIS environment after receiving research approvals from the relevant UK-CRIS trust.

\section*{Acknowledgments}

The study was funded by the National Institute for Health Research’s (NIHR) Oxford Health Biomedical Research Centre (BRC-1215-20005). This work was supported by the UK Clinical Records Interactive Search (UK-CRIS) system funded and developed by the NIHR Oxford Health BRC at Oxford Health NHS Foundation Trust and the Department of Psychiatry, University of Oxford. AK, NV, QL, ANH are funded by the MRC Pathfinder Grant (MC-PC-17215). HN was supported by the Alan Turing Institute under the EPSRC grant EP/N510129/1 and under the EPSRC EP/S026347/1. The views expressed are those of the authors and not necessarily those of the UK National Health Service, the NIHR, or the UK Department of Health. We highly appreciate the work and support of the Oxford CRIS Team: Tanya Smith, Lulu Kane, Adam Pill and Suzanne Fisher and Dr Peter Phiri of the Southern Health NHS Foundation Trust CRIS team.

\bibliographystyle{acl_natbib}
\bibliography{nlp_sig_refs}

\begin{thebibliography}{41}
\expandafter\ifx\csname natexlab\endcsname\relax\def\natexlab#1{#1}\fi

\bibitem[{Arribas(2018)}]{arribas2018derivatives}
Imanol~Perez Arribas. 2018.
\newblock Derivatives pricing using signature payoffs.
\newblock \emph{arXiv preprint arXiv:1809.09466}.

\bibitem[{Arribas et~al.(2018)Arribas, Saunders, Goodwin, and
  Lyons}]{arribas2017signature}
Imanol~Perez Arribas, Kate Saunders, Guy Goodwin, and Terry Lyons. 2018.
\newblock A signature-based machine learning model for bipolar disorder and
  borderline personality disorder.
\newblock \emph{Translational Psychiatry}, 8.

\bibitem[{Bahdanau et~al.(2014)Bahdanau, Cho, and Bengio}]{bahdanau2014neural}
Dzmitry Bahdanau, Kyunghyun Cho, and Yoshua Bengio. 2014.
\newblock Neural machine translation by jointly learning to align and
  translate.
\newblock \emph{arXiv preprint arXiv:1409.0473}.

\bibitem[{Bai et~al.(2018)Bai, Zhang, Egleston, and
  Vucetic}]{bai2018interpretable}
Tian Bai, Shanshan Zhang, Brian~L Egleston, and Slobodan Vucetic. 2018.
\newblock Interpretable representation learning for healthcare via capturing
  disease progression through time.
\newblock In \emph{Proceedings of the 24th ACM SIGKDD International Conference
  on Knowledge Discovery \& Data Mining}, pages 43--51.

\bibitem[{Chevyrev and Kormilitzin(2016)}]{chevyrev2016primer}
Ilya Chevyrev and Andrey Kormilitzin. 2016.
\newblock A primer on the signature method in machine learning.
\newblock \emph{arXiv preprint arXiv:1603.03788}.

\bibitem[{Chevyrev et~al.(2018)Chevyrev, Nanda, and
  Oberhauser}]{chevyrev2018persistence}
Ilya Chevyrev, Vidit Nanda, and Harald Oberhauser. 2018.
\newblock Persistence paths and signature features in topological data
  analysis.
\newblock \emph{IEEE transactions on pattern analysis and machine
  intelligence}, pages 1--1.

\bibitem[{Chevyrev and Oberhauser(2018)}]{chevyrev2018signature}
Ilya Chevyrev and Harald Oberhauser. 2018.
\newblock Signature moments to characterize laws of stochastic processes.
\newblock \emph{arXiv preprint arXiv:1810.10971}.

\bibitem[{Collett(2015)}]{collett2015modelling}
David Collett. 2015.
\newblock \emph{Modelling survival data in medical research}.
\newblock CRC press.

\bibitem[{Committee et~al.(2019)}]{joint2019bnf}
Joint~Formulary Committee et~al. 2019.
\newblock \emph{BNF 77 (British National Formulary) March 2019}.
\newblock Pharmaceutical Press.

\bibitem[{Gligic et~al.(2019)Gligic, Kormilitzin, Goldberg, and
  Nevado-Holgado}]{gligic2019named}
Luka Gligic, Andrey Kormilitzin, Paul Goldberg, and Alejo Nevado-Holgado. 2019.
\newblock Named entity recognition in electronic health records using transfer
  learning bootstrapped neural networks.
\newblock \emph{arXiv preprint arXiv:1901.01592}.

\bibitem[{Goodday et~al.(2020)Goodday, Kormilitzin, Vaci, Liu, Cipriani, Smith,
  and Nevado-Holgado}]{goodday2020maximizing}
SM~Goodday, A~Kormilitzin, N~Vaci, Q~Liu, A~Cipriani, T~Smith, and
  A~Nevado-Holgado. 2020.
\newblock Maximizing the use of social and behavioural information from
  secondary care mental health electronic health records.
\newblock \emph{Journal of Biomedical Informatics}, page 103429.

\bibitem[{Hambly and Lyons(2010)}]{hambly2010uniqueness}
Ben Hambly and Terry Lyons. 2010.
\newblock Uniqueness for the signature of a path of bounded variation and the
  reduced path group.
\newblock \emph{Annals of Mathematics}, pages 109--167.

\bibitem[{Harkema et~al.(2009)Harkema, Dowling, Thornblade, and
  Chapman}]{harkema2009context}
Henk Harkema, John~N Dowling, Tyler Thornblade, and Wendy~W Chapman. 2009.
\newblock Context: an algorithm for determining negation, experiencer, and
  temporal status from clinical reports.
\newblock \emph{Journal of biomedical informatics}, 42(5):839--851.

\bibitem[{Harrell et~al.(1982)Harrell, Califf, Pryor, Lee, and
  Rosati}]{harrell1982evaluating}
Frank~E Harrell, Robert~M Califf, David~B Pryor, Kerry~L Lee, and Robert~A
  Rosati. 1982.
\newblock Evaluating the yield of medical tests.
\newblock \emph{Jama}, 247(18):2543--2546.

\bibitem[{Henry et~al.(2020)Henry, Buchan, Filannino, Stubbs, and
  Uzuner}]{henry20202018}
Sam Henry, Kevin Buchan, Michele Filannino, Amber Stubbs, and Ozlem Uzuner.
  2020.
\newblock 2018 n2c2 shared task on adverse drug events and medication
  extraction in electronic health records.
\newblock \emph{Journal of the American Medical Informatics Association},
  27(1):3--12.

\bibitem[{Ishwaran et~al.(2008)Ishwaran, Kogalur, Blackstone, Lauer
  et~al.}]{ishwaran2008random}
Hemant Ishwaran, Udaya~B Kogalur, Eugene~H Blackstone, Michael~S Lauer, et~al.
  2008.
\newblock Random survival forests.
\newblock \emph{The annals of applied statistics}, 2(3):841--860.

\bibitem[{Johnson et~al.(2016)Johnson, Pollard, Shen, Li-wei, Feng, Ghassemi,
  Moody, Szolovits, Celi, and Mark}]{johnson2016mimic}
Alistair~EW Johnson, Tom~J Pollard, Lu~Shen, H~Lehman Li-wei, Mengling Feng,
  Mohammad Ghassemi, Benjamin Moody, Peter Szolovits, Leo~Anthony Celi, and
  Roger~G Mark. 2016.
\newblock Mimic-iii, a freely accessible critical care database.
\newblock \emph{Scientific data}, 3:160035.

\bibitem[{Kidger et~al.(2019)Kidger, Bonnier, Arribas, Salvi, and
  Lyons}]{kidger2019deep}
Patrick Kidger, Patric Bonnier, Imanol~Perez Arribas, Cristopher Salvi, and
  Terry Lyons. 2019.
\newblock Deep signature transforms.
\newblock In \emph{Advances in Neural Information Processing Systems}, pages
  3105--3115.

\bibitem[{Kidger and Lyons(2020)}]{kidger2020signatory}
Patrick Kidger and Terry Lyons. 2020.
\newblock Signatory: differentiable computations of the signature and
  logsignature transforms, on both cpu and gpu.
\newblock \emph{arXiv preprint arXiv:2001.00706}.

\bibitem[{Kormilitzin et~al.(2016)Kormilitzin, Saunders, Harrison, Geddes, and
  Lyons}]{kormilitzin2016application}
AB~Kormilitzin, KEA Saunders, PJ~Harrison, JR~Geddes, and TJ~Lyons. 2016.
\newblock Application of the signature method to pattern recognition in the
  cequel clinical trial.
\newblock \emph{arXiv preprint arXiv:1606.02074}.

\bibitem[{Kormilitzin et~al.(2017)Kormilitzin, Saunders, Harrison, Geddes, and
  Lyons}]{kormilitzin2017detecting}
Andrey Kormilitzin, Kate~EA Saunders, Paul~J Harrison, John~R Geddes, and Terry
  Lyons. 2017.
\newblock Detecting early signs of depressive and manic episodes in patients
  with bipolar disorder using the signature-based model.
\newblock \emph{arXiv preprint arXiv:1708.01206}.

\bibitem[{Kormilitzin et~al.(2020)Kormilitzin, Vaci, Liu, and
  Nevado-Holgado}]{kormilitzin2020med7}
Andrey Kormilitzin, Nemanja Vaci, Qiang Liu, and Alejo Nevado-Holgado. 2020.
\newblock Med7: a transferable clinical natural language processing model for
  electronic health records.
\newblock \emph{arXiv preprint arXiv:2003.01271}.

\bibitem[{Lambert and Chevret(2016)}]{lambert2016summary}
J{\'e}r{\^o}me Lambert and Sylvie Chevret. 2016.
\newblock Summary measure of discrimination in survival models based on
  cumulative/dynamic time-dependent roc curves.
\newblock \emph{Statistical methods in medical research}, 25(5):2088--2102.

\bibitem[{Lample et~al.(2016)Lample, Ballesteros, Subramanian, Kawakami, and
  Dyer}]{lample2016neural}
Guillaume Lample, Miguel Ballesteros, Sandeep Subramanian, Kazuya Kawakami, and
  Chris Dyer. 2016.
\newblock Neural architectures for named entity recognition.
\newblock \emph{arXiv preprint arXiv:1603.01360}.

\bibitem[{Liao et~al.(2019)Liao, Lyons, Yang, and Ni}]{liao2019learning}
Shujian Liao, Terry Lyons, Weixin Yang, and Hao Ni. 2019.
\newblock Learning stochastic differential equations using rnn with log
  signature features.
\newblock \emph{arXiv preprint arXiv:1908.08286}.

\bibitem[{Lyons(2014)}]{lyons2014rough}
Terry Lyons. 2014.
\newblock Rough paths, signatures and the modelling of functions on streams.
\newblock \emph{arXiv preprint arXiv:1405.4537}.

\bibitem[{Miao et~al.(2015)Miao, Cai, Zhang, and Li}]{miao2015random}
Fen Miao, Yun-Peng Cai, Yuan-Ting Zhang, and Chun-Yue Li. 2015.
\newblock Is random survival forest an alternative to cox proportional model on
  predicting cardiovascular disease?
\newblock In \emph{6TH European conference of the international federation for
  medical and biological engineering}, pages 740--743. Springer.

\bibitem[{Morrill et~al.(2020{\natexlab{a}})Morrill, Kidger, Salvi, Foster, and
  Lyons}]{morrill2020neural}
James Morrill, Patrick Kidger, Cristopher Salvi, James Foster, and Terry Lyons.
  2020{\natexlab{a}}.
\newblock Neural cdes for long time series via the log-ode method.
\newblock \emph{arXiv preprint arXiv:2009.08295}.

\bibitem[{Morrill et~al.(2019)Morrill, Kormilitzin, Nevado-Holgado,
  Swaminathan, Howison, and Lyons}]{morrill2019signature}
James Morrill, Andrey Kormilitzin, Alejo Nevado-Holgado, Sumanth Swaminathan,
  Sam Howison, and Terry Lyons. 2019.
\newblock The signature-based model for early detection of sepsis from
  electronic health records in the intensive care unit.
\newblock In \emph{2019 Computing in Cardiology (CinC)}, pages Page--1. IEEE.

\bibitem[{Morrill et~al.(2020{\natexlab{b}})Morrill, Kormilitzin,
  Nevado-Holgado, Swaminathan, Howison, and Lyons}]{morrill2020utilization}
James~H Morrill, Andrey Kormilitzin, Alejo~J Nevado-Holgado, Sumanth
  Swaminathan, Samuel~D Howison, and Terry~J Lyons. 2020{\natexlab{b}}.
\newblock Utilization of the signature method to identify the early onset of
  sepsis from multivariate physiological time series in critical care
  monitoring.
\newblock \emph{Critical Care Medicine}, 48(10):e976--e981.

\bibitem[{Pangman et~al.(2000)Pangman, Sloan, and
  Guse}]{pangman2000examination}
Verna~C Pangman, Jeff Sloan, and Lorna Guse. 2000.
\newblock An examination of psychometric properties of the mini-mental state
  examination and the standardized mini-mental state examination: implications
  for clinical practice.
\newblock \emph{Applied Nursing Research}, 13(4):209--213.

\bibitem[{Pennington et~al.(2014)Pennington, Socher, and
  Manning}]{pennington2014glove}
Jeffrey Pennington, Richard Socher, and Christopher~D Manning. 2014.
\newblock Glove: Global vectors for word representation.
\newblock In \emph{Proceedings of the 2014 conference on empirical methods in
  natural language processing (EMNLP)}, pages 1532--1543.

\bibitem[{P{\"o}lsterl et~al.(2015)P{\"o}lsterl, Navab, and
  Katouzian}]{polsterl2015fast}
Sebastian P{\"o}lsterl, Nassir Navab, and Amin Katouzian. 2015.
\newblock Fast training of support vector machines for survival analysis.
\newblock In \emph{Joint European Conference on Machine Learning and Knowledge
  Discovery in Databases}, pages 243--259. Springer.

\bibitem[{Pustejovsky and Stubbs(2012)}]{pustejovsky2012natural}
James Pustejovsky and Amber Stubbs. 2012.
\newblock \emph{Natural Language Annotation for Machine Learning: A guide to
  corpus-building for applications}.
\newblock " O'Reilly Media, Inc.".

\bibitem[{Reizenstein and Graham(2018)}]{reizenstein2018iisignature}
Jeremy Reizenstein and Benjamin Graham. 2018.
\newblock The iisignature library: efficient calculation of iterated-integral
  signatures and log signatures.
\newblock \emph{arXiv preprint arXiv:1802.08252}.

\bibitem[{Segura~Bedmar et~al.(2013)Segura~Bedmar, Mart{\'\i}nez, and
  Herrero~Zazo}]{segura2013semeval}
Isabel Segura~Bedmar, Paloma Mart{\'\i}nez, and Mar{\'\i}a Herrero~Zazo. 2013.
\newblock Semeval-2013 task 9: Extraction of drug-drug interactions from
  biomedical texts (ddiextraction 2013).
\newblock In \emph{Semeval-2013}. Association for Computational Linguistics.

\bibitem[{Vaci et~al.(2020{\natexlab{a}})Vaci, Koychev, Kim, Kormilitzin, Liu,
  Lucas, Dehghan, Nenadic, and
  Nevado-Holgado}]{vaci_koychev_kim_kormilitzin_liu_lucas_dehghan_nenadic_nevado-holgado_2020}
Nemanja Vaci, Ivan Koychev, Chi-Hun Kim, Andrey Kormilitzin, Qiang Liu,
  Christopher Lucas, Azad Dehghan, Goran Nenadic, and Alejo Nevado-Holgado.
  2020{\natexlab{a}}.
\newblock \href {https://doi.org/10.1192/bjp.2020.136} {Real-world
  effectiveness, its predictors and onset of action of cholinesterase
  inhibitors and memantine in dementia: retrospective health record study}.
\newblock \emph{The British Journal of Psychiatry}, page 1–7.

\bibitem[{Vaci et~al.(2020{\natexlab{b}})Vaci, Liu, Kormilitzin, De~Crescenzo,
  Kurtulmus, Harvey, O'Dell, Innocent, Tomlinson, Cipriani
  et~al.}]{vaci2020natural}
Nemanja Vaci, Qiang Liu, Andrey Kormilitzin, Franco De~Crescenzo, Ayse
  Kurtulmus, Jade Harvey, Bessie O'Dell, Simeon Innocent, Anneka Tomlinson,
  Andrea Cipriani, et~al. 2020{\natexlab{b}}.
\newblock Natural language processing for structuring clinical text data on
  depression using uk-cris.
\newblock \emph{Evidence-Based Mental Health}, 23(1):21--26.

\bibitem[{Xie et~al.(2017)Xie, Sun, Jin, Ni, and Lyons}]{xie2017learning}
Zecheng Xie, Zenghui Sun, Lianwen Jin, Hao Ni, and Terry Lyons. 2017.
\newblock Learning spatial-semantic context with fully convolutional recurrent
  network for online handwritten chinese text recognition.
\newblock \emph{IEEE transactions on pattern analysis and machine
  intelligence}, 40(8):1903--1917.

\bibitem[{Yang et~al.(2017)Yang, Lyons, Ni, Schmid, Jin, and
  Chang}]{yang2017leveraging}
Weixin Yang, Terry Lyons, Hao Ni, Cordelia Schmid, Lianwen Jin, and Jiawei
  Chang. 2017.
\newblock Leveraging the path signature for skeleton-based human action
  recognition.
\newblock \emph{arXiv preprint arXiv:1707.03993}.

\bibitem[{Zhang et~al.(2020)Zhang, Thadajarassiri, Sen, and
  Rundensteiner}]{zhang2020time}
Dongyu Zhang, Jidapa Thadajarassiri, Cansu Sen, and Elke Rundensteiner. 2020.
\newblock Time-aware transformer-based network for clinical notes series
  prediction.
\newblock In \emph{Machine Learning for Healthcare Conference}, pages 566--588.
  PMLR.

\end{thebibliography}

\end{document}